\documentclass{article}
\usepackage{amsfonts, amssymb, amsmath,bbm,amsthm,graphicx}
\usepackage{booktabs,array}
\usepackage[margin=3cm]{geometry}

\DeclareMathOperator*{\argmax}{arg\max}
\title{
Efficient batch-sequential Bayesian optimization \\
 with moments of truncated Gaussian vectors    
} 
\author{
S\'ebastien Marmin\footnotemark[1] \footnotemark[2] \footnotemark[3] \ , 
Cl\'ement Chevalier\footnotemark[4] \ ,
David Ginsbourger\footnotemark[1] \footnotemark[5]}

\begin{document}
\maketitle

\renewcommand{\thefootnote}{\fnsymbol{footnote}}
\footnotetext[1]{IMSV, Department of Mathematics and Statistics, University of Bern, Switzerland}
\footnotetext[2]{Institut de Math\'ematiques de Marseille (UMR7373), \'Ecole Centrale de Marseille, France}
\footnotetext[3]{Institut de Radioprotection et de S\^uret\'e Nucl\'eaire (IRSN), PSN-RES, SEMIA, LIMAR, Cadarache, 13115 Saint-Paul-l\`es-Durance, France}
\footnotetext[4]{Institute of Statistics, University of Neuch\^atel, Switzerland}
\footnotetext[5]{Uncertainty Quantification and Optimal Design group, Idiap Research Institute, Martigny, Switzerland}

\renewcommand{\thefootnote}{\arabic{footnote}}

\newcommand \zzero{\boldsymbol 0}
\newcommand \oone{\boldsymbol 1}
\renewcommand{\aa}{\boldsymbol{a}}
\newcommand{\bb}{\boldsymbol{b}}
\newcommand{\cc}{\boldsymbol{c}}
\newcommand{\ee}{\boldsymbol{e}}
\newcommand{\hh}{\boldsymbol{h}}
\newcommand{\kk}{\boldsymbol{k}}
\newcommand{\mm}{\boldsymbol{m}}
\renewcommand{\tt}{\boldsymbol{t}}
\newcommand{\uu}{\boldsymbol{u}}
\newcommand{\vv}{\boldsymbol{v}}
\newcommand{\ww}{\boldsymbol{w}}
\newcommand{\xx}{\boldsymbol{x}}
\newcommand{\yy}{\boldsymbol{y}}
\newcommand{\zz}{\boldsymbol{z}}
\renewcommand{\AA}{\boldsymbol{A}}
\newcommand{\HH}{\boldsymbol{H}}
\newcommand{\II}{\boldsymbol{I}}
\newcommand{\XX}{\boldsymbol{X}}
\newcommand{\YY}{\boldsymbol{Y}}
\newcommand{\ZZ}{\boldsymbol{Z}}
\newcommand{\aalpha}{\boldsymbol\alpha}
\newcommand{\ggamma}{\boldsymbol\gamma}
\newcommand{\mmu}{\boldsymbol\mu}
\newcommand{\Ssigma}{\boldsymbol\Sigma}

\newcommand{\Gcal}{\mathcal{G}}
\newcommand{\Mcal}{\mathcal M}
\newcommand{\Ncal}{\mathcal N}

\newcommand{\deriv}[2]{\frac{\partial #1}{\partial #2}}
\newcommand{\deri}[2]{\frac{\mathrm d #1}{\mathrm d #2}}
\newcommand{\drm}{\mathrm d}
\newcommand{\esp}{\mathbb{E}}
\newcommand{\prob}{\mathbb{P}}
\newcommand{\indic}[1]{\mathbbmss{1}_{\left\{#1\right\}}}
\newcommand \tr {\mathrm{tr}}
\newcommand \trans {^\top}
\newcommand{\integ}[4]{\int \limits_{#1}^{#2} \!\!\!\!\!\!\!~_{~^{\ldots}}\!\!\!\int \limits_{#1}^{#4} #3 }
\newcommand{\diffe}[2]{~\mathrm d #1_1\!\!\!\!~_{~^{\ldots}}\mathrm d #1_{#2} }

\newcommand{\qEI}{q\text{-}{\mathrm{EI}}}
\newcommand{\cov}{\mathrm{cov}}
\newcommand{\An}{\mathcal{A}_n}
\newcommand{\Nset}{\mathbb{N}}
\newcommand{\Rset}{\mathbb{R}}
\newcommand{\Sppset}{S_{++}}

\newtheorem{remark}{Remark}
\newtheorem{proposition}{Proposition}
\newtheorem{definition}{Definition}
\newtheorem{example}{Example}
\newtheorem{algorithm}{Algorithm}

\begin{abstract}

We deal with the efficient parallelization of Bayesian global optimization algorithms, and more specifically of those based on the expected improvement criterion and its variants. 
A closed form formula relying on multivariate Gaussian cumulative distribution functions is established for a generalized version of the 
multipoint expected improvement criterion. 
In turn, the latter relies on intermediate results that could be of independent interest concerning moments of truncated Gaussian vectors. 
The obtained expansion of the criterion enables studying its differentiability with respect to point batches and calculating the corresponding gradient in closed form. 
Furthermore, we derive fast numerical approximations of this gradient and propose efficient batch optimization strategies.   
Numerical experiments illustrate that the proposed approaches enable computational savings of between one and two order of magnitudes, hence enabling derivative-based batch-sequential acquisition function maximization to become a practically implementable and efficient standard.    
\end{abstract}

\pagestyle{myheadings}
\thispagestyle{plain}

\section{Introduction}
\label{sec:intro}

\begin{figure}
\centering
\includegraphics[scale=0.6]{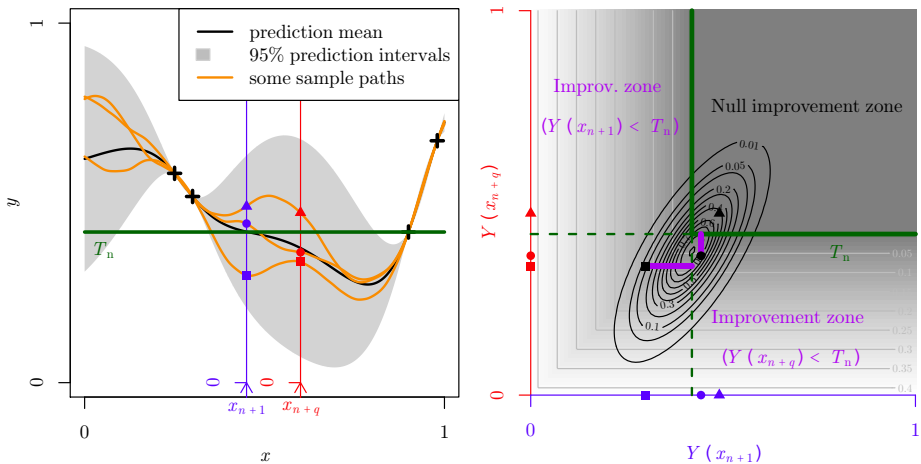}
\caption{Illustration of the principles underlying $\qEI$ for $d=1$, $n=4$, $q=2$. 
Left: Gaussian process prediction of a function $f$ from observations $\An$ (depicted by black crosses). The green horizontal line stands for $T_n$, the smallest response value from $\An$.  
Three conditional simulation draws are plotted in orange and various point symbols represent their respective values at two unobserved locations $x_{n+1}$ and $x_{n+q}$. Right: distribution of the random vector $\left(Y(x_{n+1}),Y(x_{n+q})\right)\trans$ knowing $\An$ (black contours). 
For each point symbol, the length of the purple segment represents the improvement realized by the corresponding sample path. 
The multipoint EI is the expectation of this length, or in other words, it is the integral of the improvement (grey-scale function) with respect to the conditional distribution of $\left(Y(x_{n+1}),Y(x_{n+q})\right)\trans$ knowing $\An$. 
%\textcolor{red}{Note CC : je suis partisan de montrer cette figure dans l'introduction, ou bien de ne pas la montrer du tout.} \textcolor{blue}{Note S\'eb : Moi j'aime bien la figure. Elle montre bien ce qu'est un $\qEI$ et peut-\^etre pourquoi il y a des moments de gaussiennes tronqu\'es.}
%
% Note David : ok pour la figure, mais code couleur à changer (violet pas bien lisible...)
}
\label{fig:explanation}
\end{figure}

Since their beginnings about half a century ago \cite{kushner64, zhilinskas1975single,Mockus}, Bayesian optimization algorithms have been increasingly used for derivative-free global minimization of expensive to evaluate functions. Typically assuming a continuous objective function $f:\xx\in D \subset \Rset^d \longrightarrow f(\xx) \in \Rset$, single-objective Bayesian optimization algorithms consist in sequentially evaluating $f$ at promising points under the assumption that $f$ is a sample realization (\textit{path} or \textit{trajectory}) of a random field $(Y(\xx))_{\xx \in D}$. 
Such algorithms are especially popular in the case where evaluating $f(\xx)$ requires heavy high-fidelity numerical simulations (or \textit{computer experiments}, see notably \cite{ohagan:1978:curvefitting,sac89,schonlau:phd,Jones}), where $\xx$ stands for some design parameters to be optimized over. Such expensive simulations are classically encountered in the resolution of partial differential equations from physical sciences, engineering and beyond \cite{forrester:2008:edsm}. % Refs? 
In recent years, Bayesian optimization also has attracted a lot of interest from the machine learning community \cite{Lizotte2008, Brochu.etal2009, Osborne2010, Srinivas.etal2010}, be it to optimize simulation-based objective functions \cite{Lizotte.etal2007, Tesch.etal2011, Romero.etal2013} or even to estimate tuning parameters of machine learning algorithms themselves \cite{Bartz-Beielstein.etal, Bergstra.etal2011,Snoek.etal2012}. 
In both communities, a Gaussian random field (or \textit{Gaussian Process}, GP) model is often used for $Y$, so that prior information on $f$ is taken into account through a trend function $m:D \longrightarrow \Rset$ and a covariance kernel $k:(\xx,\xx'):D\times D \longrightarrow \Rset$. Once $m$ and $k$ are specified, possibly up to some parameters to be inferred based on data, the considered GP model can be used as an instrument to locate the next evaluation point(s) via so-called infill sampling criteria, also referred to as  \textit{acquisition functions} or simply as \textit{criteria}. 
While a number of Bayesian optimization criteria have been proposed in the literature (see, e.g., \cite{jon01,frazier:2008:kgp,villemonteix:2009:iago,Srinivas.etal2010,
Contal.etal2014} and references therein), we concentrate here essentially on the \textit{Expected Improvement} (EI) criterion \cite{MockusB, Jones} and on variations thereof, with a focus on its use in synchronous batch-sequential optimization. 
Denoting by $\xx_{1}, \dots, \xx_{n} \in D$ points were $f$ is assumed to have already been evaluated and by 
$\xx_{n+1:n+q}:=(\xx_{n+1}, \dots, \xx_{n+q}) \in D^q$ a batch of candidate points where to evaluate $f$ next, the multipoint EI is defined as
\begin{equation} 
\label{qei}
EI_{n}(\xx_{n+1:n+q})=\mathbb{E}_{n} \left( 
\left(\min_{i=1,\hdots,n} Y(\xx_{i})-\min_{j=n+1,\hdots,n+q} Y(\xx_{j})\right)_+
\right),
\end{equation}
\noindent
where $\mathbb{E}_{n}$ refers to the conditional expectation knowing the event 
$\An:=\{Y(\xx_{1})=f(\xx_{1}),\dots,Y(\xx_{n})=f(\xx_{n})\}$.
%
%
%{\color{blue}[[[Proposition pour \'evoquer la figure \ref{fig:explanation} ::::}
%The random variable inside the expectation operator quantifies the progress in optimizing $f$ by evaluating at $\xx_{n+1:n+q}$. Three samples of this variable are represented in Figure~\ref{fig:explanation}.
%
One way of calculating such criterion is to rely on Monte Carlo simulations. Figure~\ref{fig:explanation} illustrates both what the criterion means and how to approach it by simulations, relying on three samples from the multivariate Gaussian distribution underlying 
Equation~\eqref{qei}. 
%
% Se referer à Frazier et al. dès ce stade?
%
Our main focus here, in contrast, is on deriving Equation~\eqref{qei} in closed form, studying the criterion's differentiability, and ultimately calculating and efficiently approximating its gradient in order to perform efficient batch optimization using derivative-based deterministic search.
%{\color{blue}]]]}
%

Now, for $q=1$, it is well known that EI can be expressed in closed form as a function of $m_{n}(\xx)=\mathbb{E}_{n}(Y_{\xx})$ and $s_{n}(\xx)=\sqrt{\operatorname{var}_{n}(Y(\xx))}$ as follows 
\begin{equation} 
\label{eiform}
EI_{n}(\xx)=s_{n}(\xx)\left( u_{n}(\xx)\Phi(u_{n}(\xx))+\varphi(u_{n}(\xx)) \right) 
\text{ if } s_{n}(\xx)\neq 0 \text{ and $0$ else}, 
\end{equation}
%where $u_{n}(\xx)=(m_{n}(\xx)-\max_{i=1,\dots,n}f(\xx_{i}))/s_{n}(\xx)$  (defined for
where $u_{n}(\xx)=(\min_{i=1,\dots,n}f(\xx_{i})-m_{n}(\xx))/s_{n}(\xx)$  (defined for $s_{n}(\xx)\neq 0$) and $\Phi, \varphi$ are the cumulative distribution function and probability density function of the standard Gaussian distribution, respectively.

When deriving Equation~\eqref{eiform}, Equation~\eqref{qei} happens (hence for $q=1$) to involve a first order moment of the truncated univariate Gaussian distribution. 
As shown in \cite{chevalier:multipointEI} and developed further here, it turns out that Equation~\eqref{qei} can %more generally 
be expanded in a similar way in the multipoint case ($q\geq 2$) relying on moments of truncated Gaussian vectors. This is essential for the open challenges tackled here of efficiently calculating and optimizing the multipoint criterion of Equation~\eqref{qei}. 

\medskip

The applied motivation for having batch-sequential EI algorithms is strong, as distributing evaluations of Bayesian optimization algorithms over several computing units allows significantly reducing wall-clock time and with the fast popularization of clouds, clusters and GPUs in recent years it is becoming always more commonplace to launch several calculations in parallel. Even at a slightly inflated price and scripting effort, reducing the total time off is often a primary goal in order to deliver conclusions involving heavy experiments, be they numerical or laboratory experiments, in studies subject to hard time limitations. 
Obviously, given its practical importance, the question of parallelizing EI algorithms and alike by selecting $q > 1$ points per iteration has been already tackled in a number of works from various disciplinary horizons (including notably \cite{Queipo.etal2006,Azimi.etal2010, Desautels.etal2012, Contal.etal2013,Gonz'alez.etal}). Here we essentially focus on approaches relying on the maximization of Equation~\eqref{qei} and related multipoint criteria.
The multipoint EI of Equation~\eqref{qei} has been defined in \cite{MockusB,schonlau:phd} and first calculated in closed form for the case $q=2$ in \cite{ginsbourger:2010:kws}. For the case $q \geq 3$, a Monte Carlo scheme and some sub-optimal batch selection strategies were proposed. Further work on Monte Carlo simulations for multipoint EI estimation can be found in 
\cite{janusevskis:2012,girdziusas2012parallel}; besides this, stochastic simulation ideas have been explored in \cite{frazier:2012} for maximizing this multipoint EI criterion via a stochastic gradient algorithm, an approach recently investigated in \cite{Wang.etal}. 
Meanwhile, a closed-form formula for the multipoint EI relying on combinations of $(q-1)$-and $q$-dimensional Gaussian cumulative distribution functions was obtained in \cite{chevalier:multipointEI}, a formula which applicability in reasonable time is however restricted to moderate $q$ (say $q\leq 10$) in the current situation.  
Building upon \cite{chevalier:multipointEI}, \cite{marmin2015differentiating} recently calculated the gradient of the multipoint EI criterion in closed form and obtained some first experimental results on (non-stochastic) gradient-based multipoint EI maximization. 

\medskip

Our aim in the present paper is to present a set of novel analytical and numerical results pertaining to the calculation, the computation, and the maximization of the multipoint EI criterion. As most of these novel results apply to a broader class of criteria, we first present in Section~\ref{sec:framework} a generalization of the multipoint EI that allows accounting for noise in conditioning observations and also  exponentiating the improvement. This generalized criterion is calculated using moments of truncated Gaussian vectors in the flavour of  \cite{chevalier:multipointEI}. 
The obtained formula is then revisited in the standard case (noise-free with an exponent set to $1$), leading to a numerical approximation of the multipoint EI with arbitrary precision and very significantly reduced computation time. Next, the $(qd)$-dimensional maximization of the multipoint EI criterion is discussed in Section~\ref{sec:discussion}, where the differentiability of the generalized criterion is studied, its analytical gradient is calculated, and further numerical approaches for fast gradient approximations with controllable accuracy are presented. 
Finally, Section~\ref{sec:application} is dedicated to numerical experiments where, in particular, a multistart derivative-based multipoint EI maximization algorithm highlighting the benefits of the considered methodological principles and the proposed fast approximations is tested and compared to baseline strategies. 

\section{Criteria in parallel Bayesian optimization}
\label{sec:framework}

\subsection{ General definition of Expected Improvement}
\label{sec:framework:definition}

Throughout this section the objective function $f$ may be observed noise-free or in noise, meaning that at some arbitrary iteration $i$ the observed value may be $f(\xx_{i})$ or  $f(\xx_{i})+\varepsilon_{i}$ where $\varepsilon_{i}$ is a realization of a zero mean Gaussian random variable with known 
(or estimated and plugged-in) variance. 
$f$ is assumed to be one realization of a random field $Y$, where $Y$ has a Gaussian random field (GRF) distribution conditionally to events of the form $\An := \{Y(\xx_1)=f(\xx_1),\hdots,Y(\xx_n)=f(\xx_1)\}$ (with conditioning on $Y(\xx_i)+\varepsilon_{i}$ in the noisy case, see for instance  \cite{picheny2013quantile}). 
This setup naturally includes the case where $Y$ is a GRF, but also the so-called Universal Kriging settings where $Y$ is the sum of a trend with an improper prior and a GRF \cite{ohagan:1978:curvefitting,
oakley:2002:bayesian}. Note that in noisy cases the $\varepsilon_{i}$'s are generally assumed to be independent (although the case of $\varepsilon_{i}$'s forming a Gaussian vector is tractable), but more essentially they are assumed independent of $Y$. 

\medskip

In batch-sequential Bayesian Optimization we are interested in computing sampling criteria $J_n$   depending on $q\geq 1$ new points 
$\xx_{n+1:n+q} = (\xx_{n+1},\hdots,\xx_{n+q})\in D^q $. At any step of corresponding synchronous parallel algorithms, the next batch of $q$ points $\xx^\star_{n+1:n+q}$ is then defined by globally maximizing $J_n$ over all possible batches:
\begin{equation}
\xx^\star_{n+1:n+q}\in \argmax_{\xx_{n+1:n+q}\in D^q} J_n(\xx_{n+1:n+q}).
\end{equation}
Values of such criteria typically depend on $\xx_{n+1:n+q}$ through the conditional distribution $Y(\xx_{1:n+q})|\An$, simplifying to $Y(\xx_{n+1:n+q})|\An$ in the noiseless context. Conditional mean and covariance functions are analytically formulated via the so-called \emph{kriging equations}, see e.g. \cite{ginsbourger:dicekrigingoptim}. 
Working out these criteria thus generally boils down to Gaussian vector calculus, which may become  intricate and quite cumbersome to implement as $q$ (or $n+q$, in noisy settings) increases.
Our considered generalized version of the multipoint EI criterion, that allows accounting for a Gaussian noise in the conditioning observations and also for an exponentiation in the definition of the improvement, is defined as:
\begin{equation}
\label{eq:qEI:definition}
EI_n(\xx_{n+1:n+q}) = \esp_n \left( 
\left(\min_{\ell=1,\hdots,n} Y(\xx_{\ell})-\min_{k=1,\hdots,q} Y(\xx_{n+k})\right)_+^\alpha
\right),
\end{equation}
where $\alpha\in \Nset\backslash\{0\}$, $\esp_n(\cdot) = \esp(\cdot | \An)$ and 
$(\cdot)_+ := \max(0,\cdot)$.
This form gathers several sampling criteria notably including $\qEI$, both in noiseless and noisy settings, and also a multipoint version of the generalized EI of \cite{schonlau:phd}.
In addition, the obtained results apply to batch-sequential versions of the Expected Quantile Improvement~\cite{picheny2013quantile} (EQI) and variations thereof, by a simply change of process from $Y$ to the quantile process.  
We will show in proposition \ref{proposition:analyticQei} that such generalized multipoint EI criteria can be formulated as a sum of moments of truncated Gaussian vectors. In the next subsection, in order to get a closed form for the generalized EI we first define these moments and derive some first analytical formulas, that might also be of relevance in further contexts.

\subsection{Preliminaries on moments of truncated Gaussian distribution}
\label{sec:framework:preliminaries}
We fix $\alpha \in \Nset\backslash\{0\}$ 
%, the so-called local-global parameter, 
and  $p=n+q$ in noisy settings or $p=q$ in noiseless settings.  
%the size of the random vector $\left.\left(Y(\xx_{1}),\ldots,Y(\xx_{n+q})\right)\right| \An$  (or $\left.\left(Y(\xx_{n+1}),\ldots,Y(\xx_{n+q})\right)\right| \An$ in noiseless setting).
\\
\begin{definition}
Let $\ZZ$ be a Gaussian vector with mean $\mm \in \Rset^p$ and covariance matrix $\Sigma \in S^{p}_{++}$, where $S^{p}_{++}$ is the cone of positive definite matrices of $\Rset^{p\times p}$. 
For all positive integer $k\le p$, we define the function $\Mcal_{k,\alpha}$ on $\Rset^p\times S^{p}_{++}$ by 
\begin{equation}
\label{eq:momentDefinition}
\Mcal_{k,\alpha} : (\mm,\Sigma) \mapsto \Mcal_{k,\alpha}(\mm,\Sigma) = \esp_n\left(Z_k^{\alpha} ~ \indic{\ZZ\le \zzero} \right),\end{equation}
\noindent
where the inequality $\ZZ \le \zzero$ is to be interpreted component-wise.
\end{definition}
%{\textsc{Notation.}}  

\medskip

\noindent
If $\ZZ$ is composed of values of a GRF at a batch of $q$ locations $\xx_{n+1:n+q}$, we use the notation 
%especially when the distribution of $\ZZ$ depends on variables $\xx_{n+1:n+q}$, 
$\Mcal_{k,\alpha}(\ZZ(\xx_{n+1:n+q})):=\Mcal_{k,\alpha}(\mm(\xx_{n+1:n+q}),\Sigma(\xx_{n+1:n+q}))$.
We obtain the moments $\Mcal_{k,\alpha}(\mm,\Sigma)$ of a truncated Gaussian distribution by an extension of Tallis' technique \cite{tallis} to any order, presented in the following proposition:

\bigskip

\begin{proposition}
\label{proposition:analyticalMoment}
The function $\Gcal: \Rset^p\times\Rset^p\times \Sppset^p 
%(\tt,{\mm,\Sigma}) 
\rightarrow \Rset$  
defined by
%of class $\mathrm C^\infty$ defined on $$:
\begin{equation}
\label{eq:momentGenerator}
\Gcal(\tt , {\mm,\Sigma})
=
 e^{\frac{1}{2}\left(\left(\tt+\Sigma^{-1}\mm\right)\trans \Sigma\left(\tt+\Sigma^{-1}\mm\right)-\mm\trans\Sigma^{-1}\mm\right)} \Phi_{p,\Sigma}\left(-\mm-\Sigma \tt\right), 
%\in \Rset, 
\end{equation}
where $\Phi_{p,\Sigma}(\cdot)$ is the cumulative distribution function of the centered $p$-variate normal distribution, 
%with non-singular covariance matrix $\Sigma$ 
is infinitely differentiable, and the moments $\Mcal_{k,\alpha}$ are given by: 
%partial derivatives of $\Gcal$ at $\tt=\zzero$:
\begin{equation}
\label{eq:generalMoment}
\Mcal_{k,\alpha} (\mm,\Sigma) = \left.\frac{\partial^{\alpha}\Gcal(\cdot , {\mm,\Sigma})}
{\partial t_k^\alpha}\right|_{\tt=\zzero}.
\end{equation}
\end{proposition}

%\bigskip

\noindent
The proof of this Proposition is given in appendix  \ref{appendix:analyticalMoment} and relies on calculating the moment generating function $\tt \rightarrow \esp\left(\exp\left(\tt\trans \ZZ\right) \indic{\ZZ\le\zzero} \right)$.
Even if an analytical formula can be obtained at any order of differentiation $\alpha$, the complexity of derivatives in equation \eqref{eq:generalMoment} increases rapidly. We give below the results for $\alpha$ equals 1 and 2.

\subsubsection*{Case $\alpha=1$} 
Differentiating $\Gcal$ with respect to $\tt$ yields:
\begin{align}
\deriv{\Gcal}{\tt}(\tt,\mm,\Sigma) = &
\exp\left({\frac{1}{2}\left(\left(\tt+\Sigma^{-1}\mm\right)\trans \Sigma\left(\tt+\Sigma^{-1}\mm\right)-\mm\trans\Sigma^{-1}\mm\right)}\right)
\times 
\nonumber\\\nonumber
&\left(
\Sigma\left(\tt+\Sigma^{-1}\mm\right) \Phi_{p,\Sigma}\left(-\mm-\Sigma \tt\right)
-% \left\langle
 \Sigma \nabla \Phi_{p,\Sigma}\left(-\mm-\Sigma \tt\right) %\right\rangle_{\Rset^{d\times d}} 
\right)
\end{align}
where $\nabla\Phi_{p,\Sigma}$ is the gradient of $\Phi_{p,\Sigma}$ (see appendix \ref{appendix:CDFgradient} for an analytical derivation).
%and $\langle \cdot, \cdot \rangle_{\Rset^{d\times d}}$ refers to the usual Frobenius scalar product over $\Rset^{d\times d}$.
Taking $\tt=\zzero$ 
% and considering the $k^\text{th}$ component we have:
in the previous equation gives
\begin{align}
\label{eq:tallisOrdre1}
\Mcal_{k,1}(\mm,\Sigma)
% &\equiv \esp\left(Z_k\indic{\ZZ\le\zzero_q}\right)= \frac{\partial \Gcal_{\ZZ}}{\partial t_k}(\zzero_q)\nonumber \\&
=m_k\Phi_{p,\Sigma}(-\mm) - \Ssigma_k\trans \nabla\Phi_{p,\Sigma}(-\mm)
%\sum_{j=1}^{p} \Sigma_{jk} \varphi_{\Sigma_{jj}}(-m_j) \Phi_{p-1,\Sigma_{|j}}(-\mm_{|j}),
\end{align}
where $\Ssigma_k$ is the $k^\text{th}$ column of $\Sigma$. 
It is shown in appendix \ref{appendix:CDFgradient} that computing 
each of the $p$ components of $\nabla\Phi_{p,\Sigma}$ 
requires to compute a multivariate CDF of the normal distribution 
in dimension $p-1$. The number of calls to this function 
for computing the first moment 
of the truncated Gaussian distribution is thus of $O(p)$.
%
% NOTE THAT 2.6 GENERALIZES 1.2
%
%\begin{romannum}
%\item $\mm_{|j}$ is the mean vector of $(\ZZ_{-j}|Z_j=0)$ ($\ZZ_{-j}$ being the random vector $\ZZ$  with the $j^\text{th}$ component removed),
%\item $\Sigma_{|j}$ is the covariance matrix of $(\ZZ_{-j}|Z_j=0)$.
%\end{romannum}

\medskip

\subsubsection*{Case $\alpha=2$} 
Similarly, differentiating $\Gcal$ twice with respect to $\tt$ %and taking $\tt=\zzero$ 
yields
\begin{equation}
\begin{split}
\label{eq:tallisOrdre2}
\Mcal_{k,2}(\mm,\Sigma)
&=(\Sigma_{kk} + m_k^2) \Phi_{p,\Sigma}(-\mm) + \Ssigma_k\trans~\nabla\nabla\trans \Phi_{p,\Sigma}(-\mm)\Ssigma_k\\
&+ 2 m_k \Mcal_{ k,1}(\mm,\Sigma).
\end{split}
\end{equation}
For readability, the detailed formula of 
$\nabla\nabla\trans \Phi_{p,\Sigma}$, the Hessian matrix 
of $\Phi_{p,\Sigma}$, is sent to Appendix \ref{appendix:CDFhessian}. 
The number of calls to the multivariate normal CDF 
is of $O(p^2)$.

\subsection{Analytic formulas for generalized \textit{{q}}-EI}
\label{sec:framework:analytic}~

The previous results obtained for the moments of the truncated 
normal distribution turn out to be of interest for computing the 
generalized $\qEI$ introduced in Equation~\eqref{eq:qEI:definition}, as 
shown by the following proposition.

\bigskip

\begin{proposition} 
\label{proposition:analyticQei}
For $\xx_{n+1:n+q}\in D^q$, the criterion $EI_n$ defined by \eqref{eq:qEI:definition} exists for all $\alpha$ and can be written as a sum of moments of truncated normal distributions
\begin{align}
EI_n(\xx_{n+1:n+q}) &= \sum_{\ell=1}^{n}\sum_{k=1}^{q} \Mcal_{n+k-1,\alpha}\left(\ZZ^{(\ell,k)}(\xx_{n+1:n+q})\right),
\label{eq:qEItransformed2}
\end{align}
with $\ZZ^{(\ell,k)}(\xx_{n+1:n+q})$ a vector of size $n+q-1$ defined, noting $Y_i := Y(\xx_i)$, by \\

\smallskip  
  $Z^{(\ell,k)}_i=\left\{\begin{matrix}
\\
\\
\\
\\
\end{matrix}\right.$
\begin{tabular}{ll}
$Y_\ell-Y_i$&$\text{if }1\le i\le \ell-1$, \\
$Y_\ell-Y_{i+1}$&$\text{if } \ell\le i\le n-1$, \\
$Y_k-Y_{i+1}$&$\text{if } n\le i\le n+q-1 \text{ and } i\neq n+k-1$, \\
$Y_k-Y_\ell$&$\text{if } i = n+k-1$.
\end{tabular}
\\ 

\smallskip

Moreover, in the noiseless case the random vector $\left(Y(\xx_{1}),\ldots,Y(\xx_{n})\right)$ becomes deterministic conditionally to $\An$.   
Denoting by $\ell_0$ the (smallest) index of the minimal observation,  i.e. $Y_{\ell_0} = \min_{\ell=1,\hdots,n} Y_{\ell}$, 
and writing $\ZZ^{(k)}(\xx_{n+1:n+q})$ the vector of the $q$ last components of $\ZZ^{({\ell_0},k)}(\xx_{n+1:n+q})$, Equation \eqref{eq:qEItransformed2} is simplified to:
\begin{align}
EI_n(\xx_{n+1:n+q}) &= \sum_{k=1}^{q} \Mcal_{k,\alpha}\left(\ZZ^{(k)}(\xx_{n+1:n+q})\right).
\label{eq:qEItransformed2simple}
\end{align}
%\end{remark}
\end{proposition}

\begin{remark}
In the rest of the article we also use the following compact notation for the $(n+q-1)-$dimensional vector $\ZZ^{(\ell,k)}(\xx_{n+1:n+q})$: 
\begin{equation}
\ZZ^{(\ell,k)}(\xx_{n+1:n+q})
=  A^{(\ell,k)}\left(Y(\xx_1),\ldots,Y(\xx_{n+q})\right)\trans,
\end{equation}
where $A^{(\ell,k)}$ is a matrix implicitly defined by $Z_{i}^{(l,k)}$ of proposition \ref{proposition:analyticQei}.
\end{remark}

\medskip

The proof of Proposition~\ref{proposition:analyticQei} is relegated to Appendix~\ref{appendix:qEI_sum_moments} for conciseness. 
Equation~\eqref{eq:qEItransformed2} highlights that the computation 
of the generalized $\qEI$ in noisy settings is challenging since 
it involves computing $nq$ different moments, each requiring 
$(n+q)^\alpha$ calls to the multivariate normal 
CDF in a dimension close to $n+q$. Even for $\alpha = 1$ and moderate $q$, the linear dependence in the number of observations $n$ makes the use 
of this criterion challenging in application. Regarding the noiseless criterion, the computation of $q$ moments is more affordable, at least for moderate $q$, but one has to keep in mind that the ultimate goal here is to perform global maximization of the considered criteria. It is 
thus important to bring further calculation speed-ups in order to perform this optimization in a reasonable time compared to the evaluation time of the objective function $f$, assumed expensive to evaluate. The next section discusses these matters and proposes faster formulas to compute both $\qEI$ and its gradient.
\section{Computing and optimizing the criteria}
\label{sec:discussion}

\subsection{Generalities}

Maximizing the $EI_n$ expressions given in Equation~\eqref{eq:qEItransformed2} (noisy settings) or \eqref{eq:qEItransformed2simple} (noiseless settings) is difficult.
These maximizations are performed with respect to a batch of $q$ points $\xx_{n+1:n+q}\in (\Rset^d)^q$, and are thus optimization problems in dimension $dq$. In this space, the objective function to be maximized, is not convex in general and has the interesting property that the $q$ points in the batch can be permuted without changing the value of $EI_n$; i.e. 
$EI_n( (\xx_{n+1},\hdots,\xx_{n+q}) ) =  
EI_n( (\xx_{n+\sigma(1)},\hdots,\xx_{n+\sigma(q)}) )
$ for any permutation $\sigma$ of $\{1,\hdots,q\}$.
With this property, one can reduce the measure of the search domain by $q!$, e.g. by imposing that the first coordinate of the $q$ points in the batch are in ascending order. 
% However, to the best of our knowledge, usual global optimization algorithms do not easily take advantage of such property. 
We will restrict our attention here to the use of multi-start gradient based local optimization algorithms acting on the whole input domain $D^q \subset \Rset^{dq}$, that do not exploit the structure of the problem but do not seem to be affected by this, at least with the chosen settings regarding the starting designs. %? 
Our contribution here will be to propose a faster formula for computing the first moments $\Mcal_{ k,1}$ previously presented, as well as their derivatives. This will yield an easier computation of both the 
generalized EI and its $dq$-dimensional gradient. Besides, a second approximate but faster formula to further reduce 
the calculation time of the gradient will be introduced. 

\subsection{Gradient of the generalized \textit{q}-EI}
\label{sec:discussion:gradient}
In this section, we extend the analytical gradient calculation of the $\qEI$ performed in \cite{marmin2015differentiating} to the case of the generalized noisy and noise-free $\qEI$, and provide in turn a more concise formula. Again, the presented formulas rely on results on moments 
of truncated Gaussian distributions.

\bigskip 

\begin{proposition}
Let $\xx_{n+1:n+q} \in D^q$ be a batch such that the conditional covariance matrix\\* $\left(\cov\left(\left.Y\left(\xx_{n+i}\right),Y\left(\xx_{n+j}\right)\right|\An\right)\right)_{1\le i,j\le q}$
is positive definite and the functions $\esp\left(\left.Y(\cdot)\right|\An\right)$ and \\* $\left(\cov\left(\left.Y(\cdot),Y(\xx_{n+j})\right|\An\right)\right)_{j=1,\ldots, q}$ are differentiable at each point 
$\xx_{n+i} \ (1\leq i\leq q)$. 
These derivatives are written $\mm'^{(i)}\in \Rset^d$ 
and $\Sigma'^{(i)} \in \Rset^{q\times d}$ respectively. 
In this setup, the $EI_n$ function of 
Equation~\eqref{eq:qEItransformed2} is differentiable 
and its derivative 
with respect to the j$^{\text{th}}$ 
coordinate of the point $\xx_{n+i}$ is 
%%$\forall i \le q, \forall j\le d,$
\begin{align}
\deriv{EI}{x_{ij}}(\xx_{n+1:n+q})= 
\sum_{\ell=1}^{n}\sum_{k=1}^{q} 
&m'^{(i)}_j{\AA^{(l,k)}_i}\trans
\deriv{\Mcal_{n+k-1,1}}{\mm}\left(\ZZ^{(\ell,k)}\right) +
\\
& \tr\left(A^{(l,k)}\Gamma'^{(i,j)}{A^{(l,k)}}\trans\deriv{\Mcal_{n+k-1,1}}{\Sigma}\left(\ZZ^{(\ell,k)}\right)  \right),\nonumber
\end{align}
where 
$\Gamma'^{(i,j)} = 
\left({{\Sigma'^{(i)}_{u,j}} \delta_{i,v} + 
{\Sigma'^{(i)}_{v,j}} \delta_{i,u}}\right)_{u,v}
\in \Rset^{q\times q}$, 
and $\delta$ is the Kronecker symbol. 
The derivatives $\deriv{\Mcal_{n+k-1,1}}{\mm}$ and 
$\deriv{\Mcal_{n+k-1,1}}{\Sigma}$ are calculated in Appendix~\ref{appendix:analyticalMomentDifferentiation}.
\end{proposition}

\bigskip

This new expansion of the gradient of the generalized EI 
as a sum of derivatives of first order moments will prove to 
be very useful thanks to formulas presented next.

\subsection{Fast numerical estimation of first order moments and their derivatives}
\label{sec:discussion:marminTrick1}
Let us now focus on the practical implementation of the closed-form formula of Equation~\eqref{eq:qEItransformed2}. We take $\alpha=1$ and note $p=n+q$ in noisy settings and $p=q$ in noiseless settings.
As mentioned before, the computation of the noisy or noiseless $\qEI$ (see, Eqs.~\eqref{eq:qEItransformed2},\eqref{eq:qEItransformed2simple}) 
requires calls to the CDF of the $p$ and $(p-1)-$variate normal 
distribution, $\Phi_p$ and $\Phi_{p-1}$. These CDFs are here computed using the Fortran algorithms of \cite{genz1992} wrapped in the mnormt R package \cite{mnormt}.  
A quick look at Eqs.~\eqref{eq:tallisOrdre1},\eqref{eq:qEItransformed2}
suggests that the noisy $\qEI$ requires $nq$ evaluations of $\Phi_{p}$ and $nq^2$ evaluations of $\Phi_{p-1}$. For the noiseless case (see, Equation~\eqref{eq:qEItransformed2simple}), the number 
of calls are divided by $n$. 
In both cases, a slight improvement can be obtained by noticing 
a symmetry which reduces the number of $\Phi_{p-1}$ calls 
from $nq^2$ (resp. $q^2$ in the noiseless case) to 
$nq(q+1)/2$ (resp. $q(q+1)/2$). 
This symmetry is justified in Appendix~\ref{sec:symmetry}.

\medskip

Despite this improvement, and even in the classical 
noiseless case, the number of $\Phi_{p-1}$ calls 
is still proportional to $q^2$.
We now give new efficient and trustworthy expansion that enables a fast and reliable approximation of first order moments of truncated Gaussian vectors $\Mcal_{k,1}$ by reducing this number of calls to $O(q)$. 

\medskip

\begin{proposition} Let $\varepsilon > 0$, and let $\ZZ$ 
\label{proposition:marminTrick}
be a Gaussian random vector with mean vector and covariance matrix $(\mm,\Sigma)\in \Rset^p\times\Sppset^p$. Then we have
\begin{align}
\Mcal_{k,1}(\mm,\Sigma) 
%&:=  \esp\left(	Z_k \indic{\ZZ\le \zzero} \right)\nonumber\\
& = \frac{1}{\varepsilon} \left( e^{m_k \varepsilon} \Phi_{p,\Sigma}(-\varepsilon\Ssigma_k-\mm) - \Phi_{p,\Sigma}(-\mm)\right ) + O(\varepsilon^2).
\label{eq:marminTrick}
\end{align}
\end{proposition}

\medskip

\begin{proof}
Let us consider the function $g_k : t\in \Rset\rightarrow e^{m_k t} \Phi_{p,\Sigma}(-\Ssigma_k t-\mm).$ This function $g_k$ is tangent 
at $t=0$ with the function $t\in \Rset\rightarrow \Gcal(t\ee_k)$, where the function $\Gcal$ is introduced in Proposition~\ref{proposition:analyticalMoment} and $\ee_k$ is the $k^\text{th}$ vector of the canonical basis. 
It follows from Proposition~\ref{proposition:analyticalMoment} that 
\begin{align*}
\Mcal_{k,1}(\mm,\Sigma)=&\left.\deriv{\Gcal}{t_k}(\tt, \mm,\Sigma)\right|_{t=0},
\end{align*}
and we obtain the announced result by Taylor expansion of $g_k$. 
%definition of the derivative.
\end{proof}
\bigskip

The obtained formula simply uses the approximation of a moment 
with finite differences of the moment generating function. 
We showed here that instead of fully computing the moment generating function, we can expand the simpler \emph{tangent} function $g_k$. For conciseness, we name here the use of this formula ``tangent moment method''. This formula thus enables approximating the first order moment $\Mcal_{k,1}$ at the cost of only two calls to $\Phi_p$. Hence, 
from Equation~\eqref{eq:qEItransformed2simple}, computing a noiseless 
$\qEI$ can be performed at the cost of $2q$ calls to $\Phi_q$.
Besides, a similar approach can be applied to approximate the gradient of $\qEI$ through faster computations of $\deriv{\Mcal_{k,1}}{\mm}$ and $\deriv{\Mcal_{k,1}}{\Sigma}$, as shown next:

\medskip

\begin{proposition}
The following equations hold:
\begin{align}
\deriv{\Mcal_{k,1}}{\mm} =& 
\Phi_{p,\Sigma}(-\mm)\ee_k - \frac{1}{\varepsilon} 
\left(e^{m_k\varepsilon}\nabla \Phi_{p,\Sigma}
{\left(-\Ssigma_k \varepsilon-\mm\right)}-\nabla \Phi_{p,\Sigma}
{\left(-\mm\right)}\right)
+ O(\varepsilon^2) 
\label{fastanalytic1}
\\
\deriv{\Mcal_{k,1}}{\Sigma} = &-\left(
\deriv{\Phi_{p,\Sigma}}{x_v}
\left(-\mm\right)\delta_{u,k} + 
\deriv{\Phi_{p,\Sigma}}{x_u}
\left(-\mm\right)\delta_{v,k}\right)_{u,v\le p} 
\label{fastanalytic2}
\\
\nonumber  & + \frac{1}{\varepsilon} 
\left(e^{m_k\varepsilon}\nabla\nabla\trans \Phi_{p,\Sigma}
{\left(-\Ssigma_k \varepsilon-\mm\right)} - 
\nabla\nabla\trans \Phi_{p,\Sigma}{\left(-\mm\right)}\right)
+ O(\varepsilon^2)
\end{align}
where $\nabla\nabla\trans \Phi_{p,\Sigma}$ is 
the Hessian matrix of $\Phi_{p,\Sigma}$ (see 
Appendix~\ref{appendix:CDFhessian} for details).
\end{proposition}

\medskip

As before, these formulas enable reducing the number of calls to the multivariate CDF by an order $q$. For the computation of $\qEI$ this number goes from $O(q^2)$ to $O(q)$. For computing its $dq$-dimensional gradient, it goes from $O(q^4)$ to $O(q^3)$. The latter complexity 
suggests restricting to moderate values of $q$ in applications. 
In the next section we present further results that enable further
reducing the complexity for numerically estimating the gradient. 

\subsection{A slightly biased but fast proxy of the gradient}~
\label{sec:discussion:marminTrick2}

The key idea to obtain further computational savings is summarized in this section. We first strategically decompose the gradient of moments as a sum of two terms.
\begin{proposition}
\label{proposition:productDerivative}
Let us consider a Gaussian
multivariate random field $\ZZ$ from $\Rset^d$ to 
$\Rset^p$. 
For $\xx\in\Rset^d$, let us denote by $\mm(\xx)$ 
and $\Sigma(\xx)$ the mean and the covariance matrix of $\ZZ(\xx)$. 
Let $\xx_a\in\Rset^d$ and assume that  %$\Sigma(\xx_a)\in\Sppset^p$ 
$\Sigma(\xx_a)$ is positive definite. 
Also, assume that the functions 
$\xx\rightarrow \mm(\xx)$, 
$\xx\rightarrow \Sigma(\xx)$ and $\xx\rightarrow \left(\cov(Z_i(\xx),Z_j(\xx_a))\right)_{i,j\le p}$ are differentiable 
at $\xx=\xx_a$. Then the following decomposition holds for $k=1,\ldots,p$.
\begin{align}
&\left.\nabla_{\xx}\left[
\Mcal_{k,\alpha}\left(\mm(\xx),\Sigma(\xx)\right)
\right]
\right|_{\xx=\xx_a}
:= 
\left.\nabla_{\xx}\left[
\esp\left(Z_k^\alpha(\xx)\indic{\ZZ(\xx)\le \zzero}
\right)\right]\right|_{\xx=\xx_a} \nonumber
\\
&=\left.\nabla_{\xx}\left[
\esp\left(Z_k^\alpha(\xx)\indic{\ZZ(\xx_a)\le \zzero}
\right)\right]\right|_{\xx=\xx_a} + 
\left.\nabla_{\xx}\left[
\esp\left(Z_k^\alpha(\xx_a)
\indic{\ZZ(\xx)\le \zzero}\right)\right]\right|_{\xx=\xx_a}.
\end{align}
\end{proposition}

\medskip

\begin{proof}
$\Sigma(\cdot)$ is continuous at $\xx_a$, so there exists a neightborhood $V_{\xx_a}$ of $\xx_a$ such that for all $\xx\in V_{\xx_a}$, $\Sigma(\xx)$ is positive definite.  Let us define on $V_{\xx_a}\times V_{\xx_a}$:\begin{equation}
g(\uu,\vv) = \esp\left(Z_k^\alpha(\uu)\indic{\ZZ(\vv)\le \zzero}\right).\nonumber
\end{equation}
Applying equation \eqref{eq:analyticalMomentProof} of appendix \ref{appendix:analyticalMoment}, for all $\uu$ and $\vv$, $g(\uu,\vv)$ is a moment generated by differentiation of the following function: 
\begin{equation}
M_{\uu,\vv}:t\rightarrow e^{\frac{1}{2}\left(\Sigma_{kk}(\uu)t^2+2tm_k(\uu)\right)}\Phi_{p,\Sigma(\vv)}\left(-\mm(\vv)-t\left(\cov(Z_k(\uu),Z_j(\vv))\right)_{j\le p}\trans\right).
\label{eq:analyticalUV}
\end{equation}
The analytical form of equation \eqref{eq:analyticalUV} and the assumed differentiability at $\xx_a$ ensure existence of partial derivatives of $g=(\uu,\vv)\rightarrow\frac{\drm^\alpha M_{\uu,\vv}}{\drm t^\alpha}(0)$ at $(\xx_a,\xx_a)$.
So to conclude,\begin{align*}
\left.\nabla_{\xx}\left[
\Mcal_{k,\alpha}\left(\mm(\xx),\Sigma(\xx)\right)
\right]
\right|_{\xx=\xx_a}=&
\left.\nabla_{\xx}\left[g(\xx,\xx)
\right]
\right|_{\xx=\xx_a}
\\=&
\left.\deriv{}{\uu}\left[g(\uu,\xx_a)
\right]
\right|_{\uu=\xx_a}+\left.\deriv{}{\vv}\left[g(\xx_a,\vv)
\right]
\right|_{\vv=\xx_a}
%\\=&\left.\nabla_{\xx}\left[g(\xx,\xx_a)\right]\right|_{\xx=\xx_a} + 
%\left.\nabla_{\xx}\left[
%g(\xx_a,\xx)\right]\right|_{\xx=\xx_a}
\end{align*}
\end{proof}
\bigskip

\begin{figure}
\includegraphics[width=\textwidth]{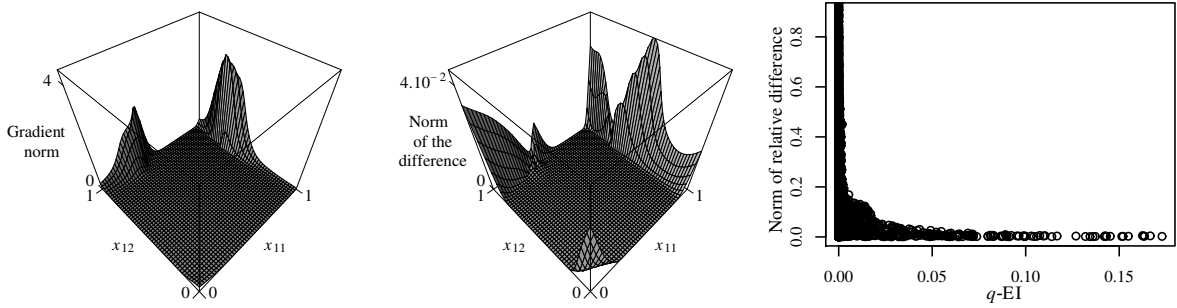}
\caption{Numerical validation of the approximation from equation \eqref{eq:nonExactSimple}, with $\alpha=1$, $q=2$, $d=2$. From left to right: 1) Norm of the $\qEI$ gradient, with respect to the first batch point (the other point is fixed in the center of $[0,1]^d$) ; 2) Norm of the difference vector between the analytical gradient and its approximation ; 3) Relative error (norm of the difference divided by the real norm) computed on 3000 random batches sampled uniformly in $[0,1]^{d\times q}$, with respect to their $\qEI$.}
\label{fig:approxComp}
\end{figure}

The latter decomposition can be interpreted as follows: infinitesimal variations of $\left(\mm(\xx),\Sigma(\xx)\right)$ around $\left(\mm(\xx_a),\Sigma(\xx_a)\right)$ modify the moments $\Mcal_{k,\alpha}\left(\mm(\xx),\Sigma(\xx)\right)$ in two ways. First, it modifies the distribution of $Z^\alpha_k(\xx)$, second it changes the distribution of the truncation $\indic{\ZZ(\xx)\le \zzero}$. For the particular case of $\qEI$, we propose to neglect this second variation. Applying this approximation to  \eqref{eq:qEItransformed2simple} gives for $X_0\in D^q$,
\begin{align}
&\left.\nabla_{\xx_{n+j}}EI(\xx_{n+1:n+q})\right|_{\xx_{n+1:n+q}=X_0}\nonumber \\ 
& = \sum_{k=1}^{q}\nabla_{\xx_{n+j}}\left. \esp\left(\left(T-Y(\xx_{n+k})\right)^\alpha\indic{A^{(k)}Y(\xx_{n+1:n+q})\le\zzero}\right)\right|_{\xx_{n+1:n+q}=X_0}\nonumber\\
&\approx\sum_{k=1}^{q}\nabla_{\xx_{n+j}}\left. \esp\left(\left(T-Y(\xx_{n+k})\right)^\alpha\indic{A^{(k)}Y(X_0)\le\zzero}\right)\right|_{\xx_{n+1:n+q}=X_0}\nonumber\\
& = - \nabla_{\xx_{n+j}}\left. \esp\left(Y(\xx_{n+j})^\alpha\indic{A^{(j)}Y(X_0)\le\zzero}\right)\right|_{\xx_{n+1:n+q}=X_0}\nonumber\\
& = -  \esp\left(\nabla_{\xx_{n+j}}
\left.Y(\xx_{n+j})^\alpha\right|_{\xx_{n+1:n+q}=X_0}
\indic{A^{(j)}Y(X_0)\le\zzero}\right), 
\label{eq:nonExactSimple}
\end{align}
where the last step is obtained by mean square 
differentiability of the process 
$\xx\rightarrow Y(\xx)^\alpha\indic{B}$, 
with $B$ an event constant with respect to $\xx$, see Appendix \ref{appendix:meanSquaredDifferentiability}. We can observe that this approximation makes a summation term disappear. The computation of this formula requires $(d+1)$ evaluations of $q$-variate Gaussian CDF. Indeed, 
Equation~\eqref{eq:nonExactSimple} indicates that each component of the gradient vector can be considered as a moment of a truncated Gaussian vector, so we can apply the results of section \ref{sec:framework}. In particular, when $\alpha=1$, applying Proposition
\ref{proposition:marminTrick}, two Gaussian CDF calls are needed for each of the $d$ components, leading to $2d$ evaluations. Besides, from Equation~\eqref{eq:marminTrick}, the second CDF call does not depend on $k$, which implies  that this term is common for every dimension. Thus the gradient of Equation~\eqref{eq:nonExactSimple} 
finally comes with $d+1$ CDF evaluations instead of $2d$. For a full gradient with respect to all $q$ points of the batch, we then need $q(d+1)$ CDF evaluations -- a substantial improvement compared to the $O(q^4)$ obtained in \cite{marmin2015differentiating} and the $O(q^3)$ obtained in the previous section. The complexities for computing moments, $\qEI$ and its gradients, expressed in terms of number of calls to the $\Phi$ function, are summarized in Table~\ref{table:tps}.
These new computational savings come at the price of a non-exact gradient calculation. A first numerical validation is represented in Figure~\ref{fig:approxComp}. On this example, we observe small ($1\times10^{-2}$) relative errors between the exact and approximate gradient of dimension $q\times d=4$ (the biggest difference vector has a norm of 0.13, compared to an exact gradient norm of 13.1). We also observe that the relative error appears to be typically smaller with higher $\qEI$, which is promising for $\qEI$ maximisations.
However, this apparently trustful but non-exact calculation naturally raises 
the question of the impact of such an approximation 
on the performances of gradient-based $\qEI$ maximization algorithms. As we will see in the next section, this proxy gradient turned out to enable quite competitive $\qEI$ maximization performances based on numerical experiments. 

\begin{table}[htbp]
\centering
\begin{tabular}{cl|cccc|c}
& 	& \multicolumn{5}{c}{Number of CDF evaluations}\\
& & $\Phi_{q-3}$ & $\Phi_{q-2}$ & $\Phi_{q-1}$ & $\Phi_{q}$ &Total\\ 
  \hline
$\Mcal_{k,1}$ & analytic      & $  $ & $  $ & $  q $ & $1$      & $O(q)$\\ 
& tangent moment & $  $ & $  $ & $   $ & $2$      & $2$\\ 
$\mathrm{EI}$& analytic            & $  $ & $  $ & 
\footnotesize
$ {q+1\choose 2}$
%$ q(q+1)/2$ 
& $q$      & $O(q^2)$\\
& tangent moment & $  $ & $  $ & $   $ & $2q$    & $O(q)$\\
\hline
$\nabla\Mcal_{k,1}$& analytic                 &  
$3{q\choose 3}$
& 
$3{q\choose 2}$
& $2q$&$1$& $O(q^3)$\\
& tangent moment    &  $ $  & 
$2{q\choose 2}$  
& $2q$& $2$ & $O(q^2)$\\
& proxy & $  $ & $  $ & $  $ & $d+1$      & $O(d)$\\
$\nabla\mathrm{EI}$ & analytic        &  
\footnotesize
$6{q+1\choose 4}$
& 
\footnotesize
$3{q+1\choose 3}$ &
\footnotesize$(3q^2+q)/2$ 
& $q$ & $O(q^4)$
\\
& tangent moment & $  $ & $q^2(q-1)$ & $ 2q^2 $ &$2q$   & $O(q^3)$\\
& proxy& $  $ & $  $ & $   $ & $q(d+1)$    & $O(qd)$\end{tabular}
\vspace{0.3cm}
\caption{In noiseless settings, total number of calls to the CDF of the multivariate Gaussian distribution for computing $\Mcal_{k,1}$, $\qEI$, their gradients and their approximations, depending on $q$ and $d$. For $\qEI$ in noisy setting, replace $q$ by $p=n+q$ and multiply each number of calls by $n$.}
\label{table:tps}
\end{table}

\section{Application}
\label{sec:application}
The goal of this section is to illustrate the usability of the proposed gradient-based $\qEI$ maximization schemes and in particular the improvements brought by the fast formulas detailed in the previous sections. The relevance of using sequential sampling strategies based on the $\qEI$ maximization has already been investigated before (see, 
\cite{chevalier:multipointEI,Wang.etal,marmin2015differentiating}) 
and all these articles pointed out the importance of calculation speed which often limits the use of $\qEI$ based strategies to moderate $q$.
We do not aim again at proving the performance of $\qEI$ based sequential strategies. Instead we aim at illustrating the gain, in computation time, brought by the fast formulas and show that using the approximate gradient obtained in Equation~\eqref{eq:nonExactSimple} does not impair the ability to find batches with (close to) maximal $\qEI$.

\subsection{Objective function and pure calculation speed}

The objective function is the so-called Borehole function  \cite{harper:sensitivity:1983}. It has been previously used for testing methods using a surrogate model
\cite{worley:deterministic:1987,gramacy:gaussian:2012}. 
The function computes a rate of water flow, $\phi$, through 
a borehole. The problem is described by $d=8$ input variables, 
$r_w \in [0.05, 0.15]$, $r \in [100, 50000]$, 
$T_u \in [63070, 115600]$, $H_u \in [990, 1110]$, 
$T_l \in [63.1, 116]$, $H_l \in [700, 820]$, 
$L \in [1120, 1680]$, $K_w \in [1500, 15 000]$ 
and is given below
%It has $d=8$ scalar input variables, denoted by 
%$(r_w,r,T_u,H_u,T_l,H_l,L,K_w)$ and computes the following:
\begin{equation}
\phi=\frac{2  \pi  T_u (H_u-H_l)}{\ln\left(\frac{r}{r_w}\right)  \left(1+ \frac{2LT_u}{  \left(\ln\left(\frac{r}{r_w}\right)r_w^2 K_w\right)}+\frac{T_u}{T_l}\right)}.
\end{equation}
Here, the objective function $f$ is obtained by rescaling $\phi$ on the 
input domain $D = [0,1]^8$. An analytical study of variations shows that there is a unique global minimum at $\xx^*=(0,1,0,0,0,1,1,0)\trans$, 
with $f(\xx^*)\approx 1.1918$.

\begin{figure}
\centering
\includegraphics[width=\textwidth]{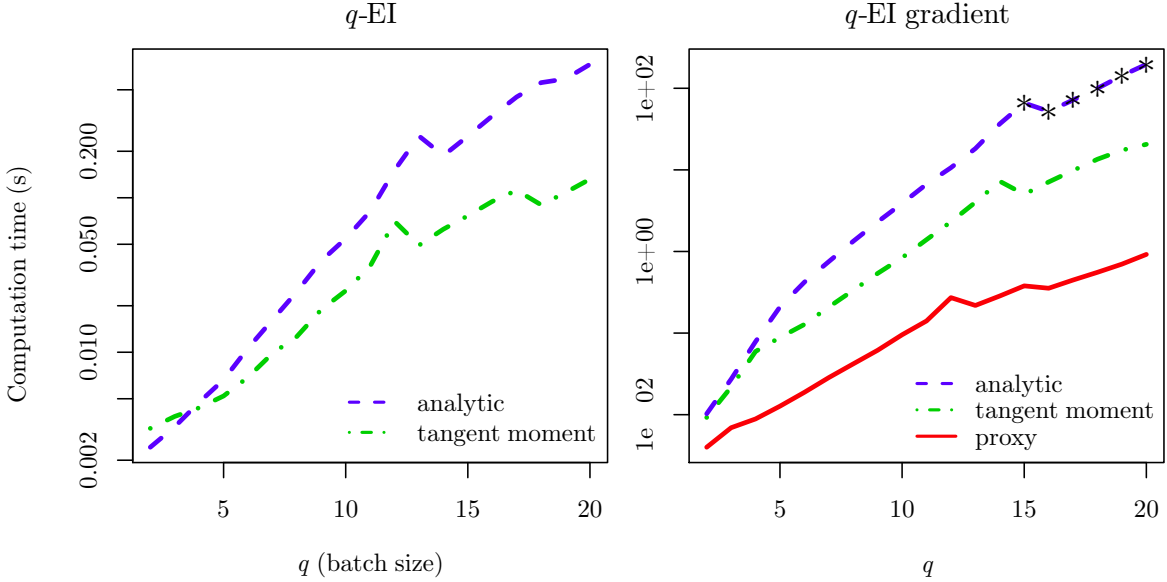}
\caption{Computation times for  $\qEI$ or its gradient as a function of
the batch size $q$ (logarithmic scale). We take an averaged computation time over $1000$ batches (except for points marked with a $*$, averaged over $150$ batches). 
}
\label{fig:computationTime}
\end{figure}

Before using sequential strategies to minimize $f$, we look at empirical 
computation times for evaluating $\qEI$ and its gradient as a function of 
the batch size $q$. For the computations, the so-called ``analytic'' method relies on the state of the art formulas 
of \cite{chevalier:multipointEI,marmin2015differentiating} with a number 
of calls to the multivariate normal cdf of respectively $O(q^2)$ and 
$O(q^4)$. The ``tangent moment'' method uses our formula for moment calculation to yield $\qEI$ and its gradient (see, Equations~\eqref{eq:marminTrick} and \eqref{fastanalytic1},\eqref{fastanalytic2}). Finally, for computing the gradient only, the ``proxy'' method relies on Equation~\eqref{eq:nonExactSimple}.

Figure~\ref{fig:computationTime} exhibits computation times averaged  
over $1000$ batches drawn uniformly. The Gaussian process model is based 
on an initial design of $n_0 = 10 d = 80$ points drawn from a optimum-LHS procedure \cite{kenny2000algorithmic}.
We use the $\text{Mat\'ern}~(\nu = 3/2)$  tensor product covariance function and estimate the hyperparameters by maximum likelihood using the 
DiceKriging R package \cite{D.Ginsbourger.etal2015}.
Figure~\ref{fig:computationTime} shows significant computational savings.
For instance with $q=8$, one gradient computation takes respectively $0.04s$, $0.33s$ and $1.33s$ using respectively the proxy, the tangent moment and analytic methods. Since the complexity for computing a gradient with the proxy is of $O(qd)$ against $O(q^3)$ and $O(q^4)$ for the two 
other methods, the computational savings of the proxy tend to increase with $q$. It should also be noted that these savings will be larger with decreasing domain dimension $d$. If we look at $\qEI$ computations, the tangent moment method is $3.3$ times faster than the analytic one when $q = 8$ and $6.5$ times faster when $q = 20$; thanks to an $O(q)$ complexity against $O(q^2)$. 
%:  respectively $0.92$, $20.56$ and $196.49$ seconds for $q=20$.

\subsection{Experimental setup: sequential minimization strategies}

We now perform a total of $50$ minimizations of $f$, each using an initial design of experiments of $n_0 = 80$ points  drawn from an optimum-LHS procedure with a different seed. Three different batch-sequential strategies are investigated. 

The first one -- serving as a benchmark -- is a variation of the ``Constant Liar Mix'' heuristic \cite{chevalier:multipointEI,Wang.etal} where, at each iteration, the batch of size $q$ is chosen among several batches obtained from the Constant Liar heuristic \cite{ginsbourger:2010:kws} with different lie levels. We use $7$ lie levels fixed to the current maximum observation the current 
minimum observation, and the $2.5\%,10\%,50\%,90\%, 97.5\%$ quantiles of the conditional distribution of the point selected in the batch. 
A total of $7$ batches are proposed at each iteration and the CL-mix heuristic picks the one with maximum $\qEI$.

The two other strategies considered here rely on pure $\qEI$ maximization 
using a multistart BFGS algorithm with a stopping criterion of precision 2.2$\times 10^{-7}$ (parameter \texttt{control\$factr} of the R function \texttt{optim} \cite{r2008r}). The gradients involved in the optimization are computed either with the tangent moment formula or the proxy. For the gradient-based $\qEI$ maximization, we use a total of $10$ starting batches obtained, again, using a Constant Liar heuristic with random lies sampled from the conditional distribution at the selected point. Finally we use two different batch sizes. When $q = 8$ we run a total of $10$ iterations and when $q = 4$ we run $20$ iterations. The hyperparameters of the GP model are re-estimated at each iteration after having incorporated the new observations. 

\subsection{First {\textit{q}}-EI maximization}

We first compare the performances, in terms of $\qEI$, of the multistart BFGS algorithm when the proxy gradient and the tangent moment methods are used. Table~\ref{tab:qEIcompare} compares the results at iteration $1$ for these two methods and the CL-mix strategy. The results are averaged over the 50 initial designs. 
\begin{table}[htbp]
\centering
\begin{tabular}{lcc}
&$q=4$&$q=8$\\
%fast analytic & 12.42067 (12.25871)&18.17365 (17.56695)\\               
%cheap approx.& 12.45665 (12.23480)&15.35287 (15.11990)\\            
%CL-mix &11.80285 (11.60988) &14.34242 (13.91853)     
tangent moment &12.45 (22.6 s)& 15.35 (700.2 s)\\
proxy&12.46 (14.3 s)& 15.35 (127.0 s)\\
CL-mix &11.80 (7.7 s)& 14.34 (15.6 s)
%fast analytic & 12.4 (12.3)&18.2 (17.6)\\               
%cheap approx.& 12.5 (12.2)&15.4 (15.1)\\            
%CL-mix &11.8 (11.6) &14.3 (13.9)           
\end{tabular}
\caption{Average $\qEI$ value of the optimal batches found for each 
of the $50$ initial designs. The numbers between brackets 
are the average computation times.}
\label{tab:qEIcompare}
\end{table}

As expected, the CL-mix heuristic yields batches with lower $\qEI$ than the strategies directly maximizing $\qEI$. Also, for both $q=4$ and $q=8$, the two $\qEI$ based methods have the same performance, which stresses out 
the relevance of the proxy method since the latter is about 1.6 times 
faster when $q=4$ and 5.5 faster when $q=8$.

\subsection{Several \textit{q}-EI maximization steps}

\begin{figure}
\centering
\includegraphics[width=14cm, clip=true, trim=0 1 0 8]{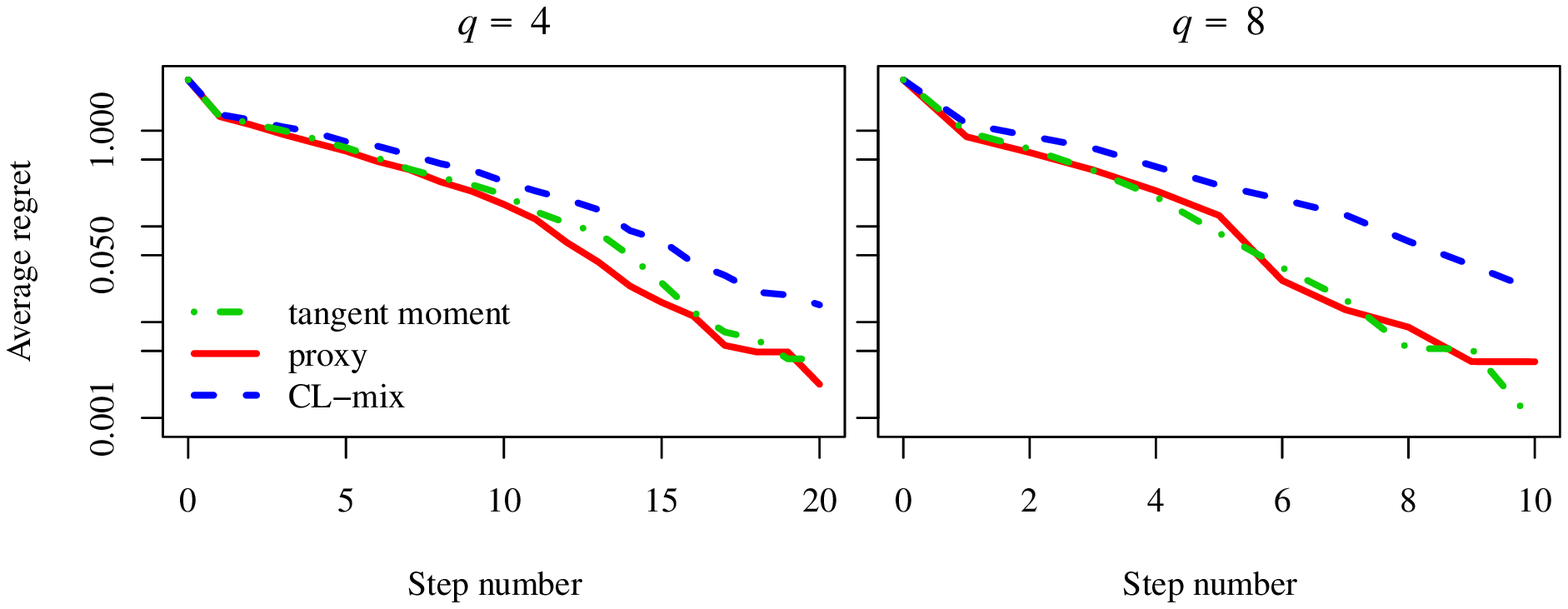}\\
\includegraphics[width=14cm, clip=true, trim=0 30 0 0]{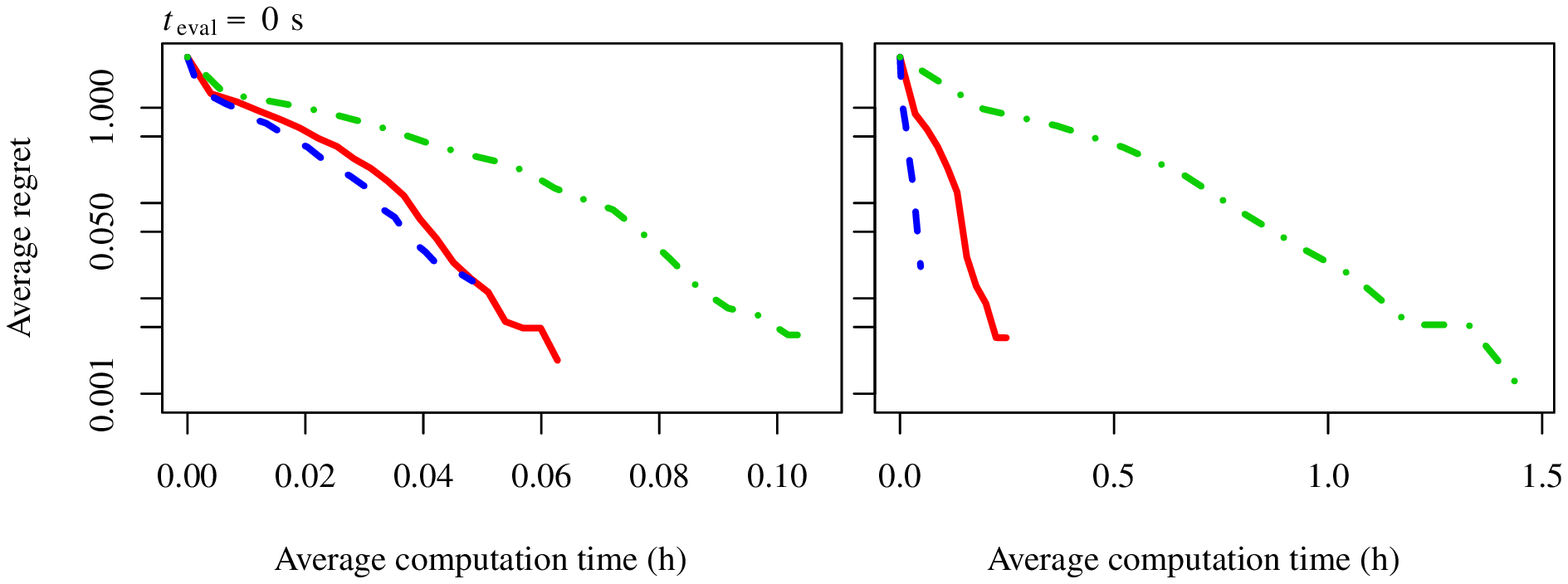}\\
\includegraphics[width=14cm, clip=true, trim=0 30 0 0]{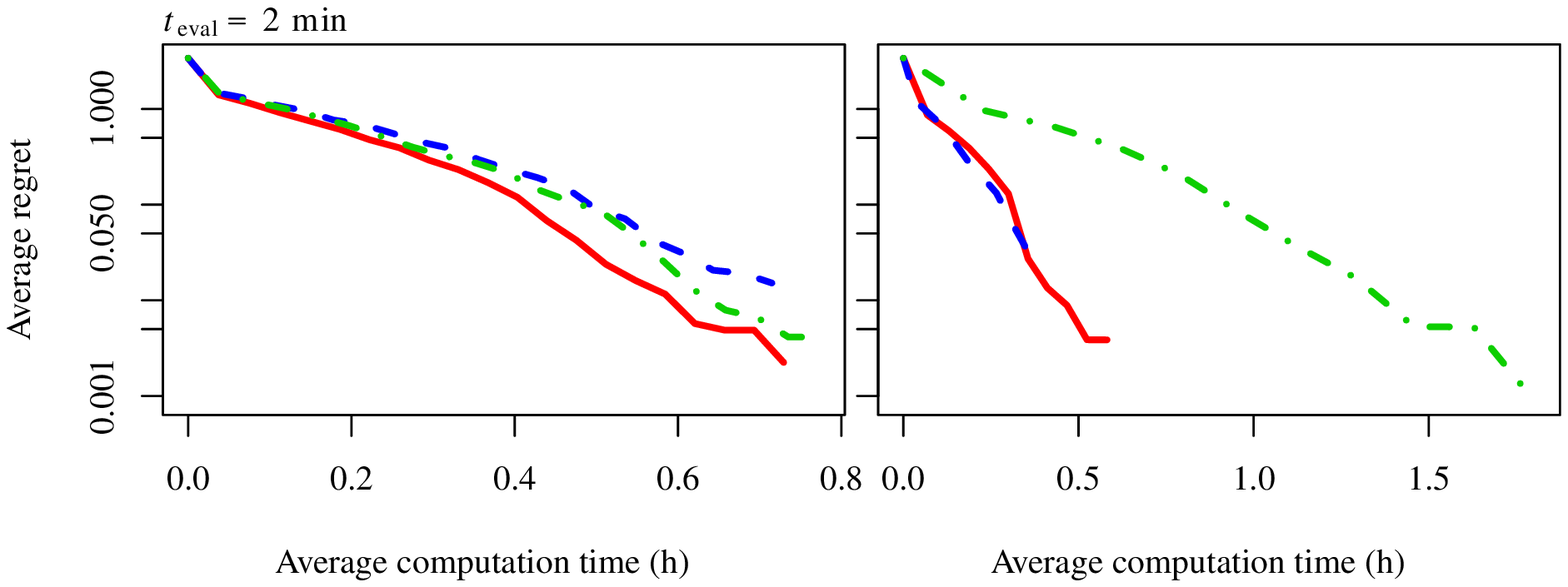}\\
\includegraphics[width=14cm, clip=true, trim=0 1 0 0]{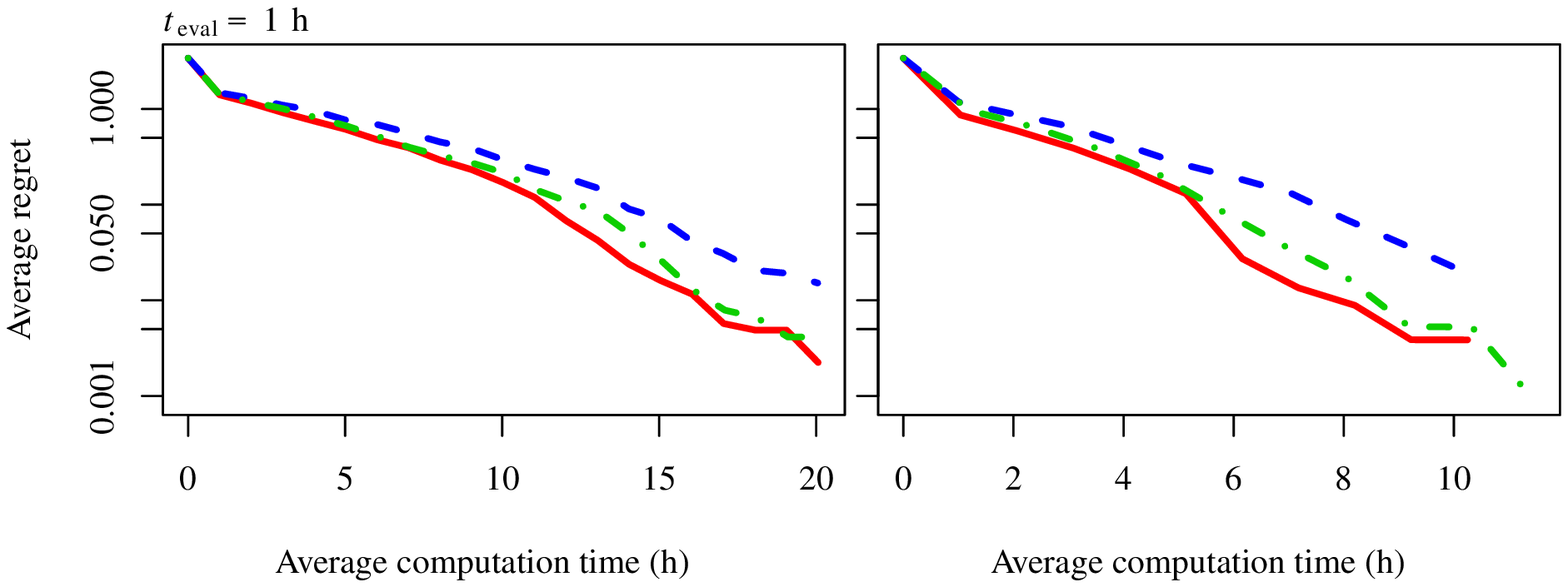}
\caption{Log-scaled average regret of the three considered optimization strategies as a function of the iteration number (row $1$) and the total computation time (rows $2,3,4$) assuming that the computation times of $f$, $t_\text{eval}$, are respectively $0$ seconds, $2$ minutes and $1$ hour. Experiments are performed with $q=4$ (left column) and $q=8$ (right column).
}
\label{fig:steps}
\end{figure}

We now compare the performances of the different $\qEI$ maximization 
approaches after multiple batch evaluations. Figure~\ref{fig:steps} 
displays the average regret as a function of the 
iteration number (first row) and the total computation time 
(i.e. the time to evaluate $f$ and find the next batch to evaluate) 
assuming respectively that the computation time of $f$ is $0$ seconds (i.e. instantaneous), two minutes and one hour (rows $2,3,4$ respectively). 
Looking at the performances as a function of the iteration number 
(first row on Figure~\ref{fig:steps}), 
the CL-mix heuristic, which samples a batch with lower $\qEI$ at each step, leads in average to a slower convergence than the two other methods, 
for both $q=4$ and $q=8$. In contrast, the two strategies based on $\qEI$ maximization have similar performances.

However, these conclusions do not hold when the regret is plotted as a 
function of the total computation time (rows $2,3,4$ on Figure~\ref{fig:steps}). First, when the computation time $t_\text{eval}$ of $f$ is null (row $2$) it is clear that $\qEI$-based sequential strategies are not adapted since they are too expensive. In this case, 
the CL-mix heuristic performs better and some other optimization strategies which are not metamodel-based would probably be more relevant. Second, when $f$ is moderately expensive (i.e. $t_\text{eval} = 2$ minutes), the proxy method and CL-mix have comparable performances when $q = 8$, but the proxy outperforms when $q=4$. Besides, the proxy shows a much faster convergence than the tangent moment method when $q = 8$.
The use of $\qEI$ based strategies thus becomes relevant when $t_\text{eval}$ is larger than a few minutes, if the proxy is used. 
Finally, when $t_\text{eval}$ is equal to one hour, the use of $\qEI$ based strategies is particularly recommended. In that case the relative 
improvement of the proxy compared to the tangent moment method tends to naturally vanish because of the long computation time of $f$. When $f$ is extremely expensive to compute, using the proxy is thus not essential. However, since it does not impair the ability to find a batch with large $\qEI$ we still recommend to use it, especially when $q$ is large.

\section*{Conclusion}

In this article we provide a closed-form expression of generalized $q$-points Expected Improvement criterion for batch-sequential Bayesian global optimization. 
An interpretation based on moments of truncated Gaussian vectors yields fast $\qEI$ formulas with arbitrary precision.  
Furthermore an new approximation for the gradient is shown to be even faster while preserving ability to find batches close to maximal $\qEI$. 
As the use of these strategies was previously considered cumbersome from a dozen of batch points, these formulas happen to be of particular interest to run $\qEI$ based batch-sequential strategies for larger batch sizes. 
We show that these methods are implementable and efficient on a classic $8$-dimensional test case. Additionally, some of the intermediate results established here might be of interest for other research questions involving moments of truncated Gaussian vectors and their gradients. Perspectives include deriving second order derivatives of $\qEI$ and fast numerical estimates thereof. Also, we aim at improving the sampling of initial batches in multistart derivative-based $\qEI$ maximization.

\medskip

\textbf{Acknowledgements}: Part of this work has been conducted within the frame of the ReDice consortium, gathering industrial (CEA, EDF, IFPEN, IRSN, Renault) and academic (\'Ecole des Mines de Saint-\'Etienne, INRIA, Universit\"at Bern) partners around advanced methods for computer experiments.

\appendix 

\section{Differentiating multivariate Gaussian CDF}

We consider the CDF dimension $p\ge 2$. We use the convention $\Phi_0=1$.
\subsection{Gradient}
\label{appendix:CDFgradient}

Using the following identity, derived from conditional distributions of a Gaussian vector,
\begin{equation}
\forall i =1,\ldots,p, ~\varphi_{p,\Sigma}\left(\xx\right) = \varphi_{1,\Sigma_{ii}}\left(x_i\right)\varphi_{p-1,\Sigma_{|_i}}\left(\xx_{-i}-\mm_{|i,x_i}\right),\nonumber
\end{equation}
 with $\mm_{|i,u}= \frac{u}{\Sigma_{ii}}\Ssigma_{-i,i}$ and $\Sigma_{|i} = \Sigma_{-i,-i}- \frac{1}{\Sigma_{ii}}\Ssigma_{-i,i}\Ssigma_{-i,i}\trans$, we reformulate the integral of the Gaussian CDF:
\begin{equation}
\forall i =1,\ldots,p, \Phi_{p,\Sigma}\left(\xx\right) = \int\limits_{-\infty}^{x_i} \varphi_{1,\Sigma_{ii}}\left(u_i\right)\Phi_{p-1,\Sigma_{|i}}\left(\xx_{-i}-\mm_{|i,u_i}\right)\drm u_i.\nonumber
\end{equation}
Here indexed minus symbols, e.g. in $\Ssigma_{-i,i}$, refer to exclusions of a line or a column.

Finally we have
\begin{equation}
\nabla \Phi_{p,\Sigma}\left(\xx\right) = \left(\varphi_{1,\Sigma_{ii}}\left(x_i\right)\Phi_{p-1,\Sigma{|i}}\left(\xx_{-i}-\mm_{|i,x_i}\right)\right)_{i=1,\ldots, p}.
\label{eq:CDFgradient}
\end{equation}

\subsection{Hessian}
\label{appendix:CDFhessian}

As for the computation of the gradient, we write \\$\forall i, j =1,\ldots,p, i\neq j,$
\begin{align*}
 \Phi_{p,\Sigma}\left(\xx\right) = \int\limits_{-\infty}^{x_i}\int\limits_{-\infty}^{x_j} \varphi_{2,\Sigma_{ij,ij}}\left(\left[{}^{u_i}_{u_j^{}}\right]\right)\Phi_{p-2,\Sigma_{|ij}}\left(\xx_{-\{i,j\}}-\mm_{|(i,j),(u_i,u_j)}\right)\drm u_j\drm u_i,\nonumber
\end{align*}
 with   
$\mm_{|(i,j),(u,u')}=\Sigma_{-\{ij\},ij}\Sigma_{ij,ij}^{-1}\left[\begin{matrix}u\\u'\end{matrix}\right]$ and
$\Sigma_{|ij}=\Sigma_{-\{ij\},-\{ij\}}  - \Sigma_{-\{ij\},ij}\Sigma_{ij,ij}^{-1}\Sigma_{-\{ij\},ij}\trans  $.

So $\forall i, j =1,\ldots,p, i\neq j,$
\begin{align*}
 \deriv{^2\Phi_q}{x_i\partial x_j}\left(\xx\right)=  \varphi_{2,\Sigma_{ij,ij}}\left(\left[{}^{x_i}_{x_j^{}}\right]\right)~ \Phi_{p-2,\Sigma_{|ij}}(\xx_{-\{i,j\}}-\mm_{|(i,j),(x_i,x_j)}).\nonumber
 \end{align*}
 
 When $i=j$, the differentiation of equation \eqref{eq:CDFgradient} gives,
\begin{align*}
\deriv{^2\Phi_q}{x_i^2}\left(\xx\right)= -\frac{1}{\Sigma_{ii}}\left(x_i\deriv{\Phi_{p,\Sigma}}{x_i}(\xx)+\sum_{\substack{j=1\\j\neq i}}^p \Sigma_{ij}\deriv{^2\Phi_{p,\Sigma}}{x_i\partial x_j}\left(\xx\right) \right).
\end{align*}

\section{Moments of truncated multivariate Gaussian distribution}

\subsection{Analytical formula (propositions \ref{proposition:analyticalMoment} and \ref{proposition:productDerivative})}\label{appendix:analyticalMoment}

We see here why we can derive an analytical formula of $\Mcal_{k,\alpha}(\mm,\Sigma)$, with $k\le s \in \Nset\backslash\{0\}$, $\mm\in\Rset^s$ and $\Sigma\in\Sppset^s$, by differentiating $\Gcal$, defined in equation \eqref{eq:momentGenerator}. It is known, see e.g. \cite{cressie}, that moments can be obtained differentiating the moment generating function $G_{\mm,\Sigma,s}$:
\begin{align*}
\Mcal_{k,\alpha}(\mm,\Sigma) = \frac{\partial^{\alpha} G_{\mm,\Sigma,s}}{\partial t_k^\alpha}(\zzero)\nonumber,
\end{align*}
with, for $r\in\{1,\ldots,s\}$, $G_{\mm,\Sigma,r}: \tt \rightarrow \esp\left(\exp\left(\tt\trans \ZZ\right) \indic{(Z_1,\ldots,Z_r)\trans\le\zzero} \right)$, $\ZZ\sim\Ncal\left(\mm,\Sigma\right)$. We derive now an analytical formula for $G_{\mm,\Sigma,r}$. As needed in Proposition~\ref{proposition:productDerivative}, we derive an analytical formula  for any $r$, and not only for $r=s$.
\begin{align}
\forall \tt \in \Rset^s,&\nonumber\\
G_{\mm,\Sigma,r}(\tt) &= \overbrace{\integ{-\infty}{0}{}{0}}^{r\text{ times}}\!\integ{-\infty}{\infty}{\exp{(\tt\trans\zz)}\varphi_\Sigma\left(\zz-\mm\right)}{\infty} \diffe{z}{s}\nonumber\\
&= \varphi_\Sigma(\zzero) \integ{-\infty}{0}{}{0}\!\integ{-\infty}{\infty}{\exp{\left(-\frac{1}{2}\left(\left(\zz-\mm\right)\trans\Sigma^{-1}\left(\zz-\mm\right)-2\tt\trans\zz\right)\right)}}{\infty} \mathrm d\zz\nonumber\\
&=  e^{-\frac{1}{2}\left(-\left(\tt+\Sigma^{-1}\mm\right)\trans \Sigma\left(\tt+\Sigma^{-1}\mm\right)+\mm\trans\Sigma^{-1}\mm\right)} \nonumber\\
& ~~~~~~\varphi_\Sigma(\zzero)\integ{-\infty}{0}{}{0}\!\integ{-\infty}{\infty}{e^{-\frac{1}{2}\left(\left(\zz-\mm-\Sigma\tt\right)\trans\Sigma^{-1}\left(\zz-\mm-\Sigma\tt\right)\trans\right)}}{\infty} \mathrm d\zz\nonumber\\
&=e^{\frac{1}{2}\left(\left(\tt+\Sigma^{-1}\mm\right)\trans \Sigma\left(\tt+\Sigma^{-1}\mm\right)-\mm\trans\Sigma^{-1}\mm\right)} \Phi_{r,\left(\Sigma_{ij}\right)_{i,j\le r}}\left(-\mm-\left(\Sigma_{ij}\right)_{i\le r,j\le s} \tt\right).
\label{eq:analyticalMomentProof}
\end{align}

In the frame of the proof of Proposition \ref{proposition:productDerivative},
\begin{itemize}
\item  if $\Sigma_k(\uu,\vv)$, the covariance matrix of $(\ZZ(\vv)\trans,Z_k(\uu))\trans$, is positive definite, we take
\begin{equation}
M_{\uu,\vv}=t\rightarrow G_{\left(\mm(\vv),m_k(\uu)\right),\Sigma_k(\uu,\vv),p}((0,\ldots,0,t)\trans),\nonumber
\end{equation}
\item else, as $\Sigma(\vv)$ is definite positive, there exists only one index $k_0$ such as $Z_k(\uu)=Z_{k_0}(\vv)$ almost surely (for example $k=k_0$ when $\uu=\vv$), and we have
\begin{equation}
M_{\uu,\vv}=t\rightarrow G_{\left(\mm(\vv)\right),\Sigma(\vv),p}((0,\underset{\substack{\uparrow\\k_0^\text{th}\text{ position}}}{\ldots,t,\ldots,}0)\trans).\nonumber
\end{equation}
In both cases, equation \eqref{eq:analyticalMomentProof} leads to equation \eqref{eq:analyticalUV}.
\end{itemize}

\subsection{Differentiation with respect to mean and covariance}
\label{appendix:analyticalMomentDifferentiation}

We differentiate here the equation \eqref{eq:tallisOrdre1} with respect to $\mm$ and $\Sigma$.
\paragraph{With respect to the mean $\mm$}
\begin{align}
\deriv{\Mcal_{k,1}}{\mm}(\mm,\Sigma)
=\Phi_{p,\Sigma}(-\mm)\ee_k-m_k\nabla\Phi_{p,\Sigma}(-\mm) + \nabla \nabla\trans\Phi_{p,\Sigma}(-\mm)\Ssigma_k.
\end{align}
\paragraph{With respect to the covariance $\Sigma$}
\begin{align}
\deriv{\Mcal_{k,1}}{\Sigma}(\mm,\Sigma)
=m_k\deriv{}{\Sigma}\Phi_{p,\Sigma}(-\mm) - \sum_{i=1}^p&\varphi_{\Sigma_{ii}}\left(-m_i\right)\Phi_{p-1,\Sigma|i}(-\mm_{|i})E^{(k,i)}\nonumber\\
+&\Sigma_{ki} \deriv{}{\Sigma_{ii}}\varphi_{\Sigma_{ii}}\left(-m_i\right)\Phi_{p-1,\Sigma|i}(-\mm_{|i})E^{(i,i)} \nonumber\\
+&\Sigma_{ki}\varphi_{\Sigma_{ii}}\left(-m_i\right)\deriv{}{\Sigma}\Phi_{p-1,\Sigma|i}(-\mm_{|i}).
\end{align}
with $\mm|_i=\mm_{-i}- \frac{m_i}{\Sigma_{ii}}\Ssigma_{-i,i}$ and $\Sigma|_i=\Sigma_{-i,-i}- \frac{1}{\Sigma_{ii}}\Ssigma_{-i,i}\Ssigma_{-i,i}\trans$. Writting $\drm_{\Sigma}\left[\mm|_{i}\right]$ $\drm_{\Sigma}\left[\Sigma|_{i}\right]$ the differential of the functions $\Sigma\rightarrow \mm|_{i}$ and $\Sigma\rightarrow \Sigma|_{i}$, we have:
\begin{align}
\drm_{\Sigma}\left[\mm|_{i}\right](H) &= \frac{m_i}{\Sigma_{ii}}\HH_{-i,i}\\
\drm_{\Sigma}\left[\Sigma|_{i}\right](H) &= H_{-i,-i}+\frac{H_{ii}}{\Sigma_{ii}^2}\Ssigma_{-i,i}\Ssigma_{-i,i}\trans-\frac{2}{\Sigma_{ii}} \HH_{-i,i}\Ssigma_{-i,i}\trans
\end{align}
\begin{align*}
\deriv{}{\Sigma}\Phi_{p-1,\Sigma|i}(-\mm_{|i})= \sum_{r=1}^p\sum_{s=1}^p&\left(-\drm_{\Sigma}\left[\mm|_{i}\right](E^{(r,s)}).\nabla \Phi_{p-1,\Sigma|_i}(-\mm|_i)\right.\\ &\left.+\tr\left(\deriv{}{\Gamma}\Phi_{p-1,\Sigma{|i}}\left(-\mm_{|i}\right). \drm_{\Sigma}\left[\Sigma_{|{i}}\right]\left(E^{(r,s)}\right)\right)\right)E^{(r,s)}
\end{align*}
with:\begin{itemize}
\item $E^{(r,s)}=(\delta_{ij})_{i,j=1,\ldots,p}$,
\item $\deriv{}{\Gamma}\Phi_{p-1,\Sigma_{|i}}(-\mm_{|i})$ the derivative of $\Gamma\rightarrow \Phi_{p-1,\Gamma}(-\mm_{|i})$ evaluated at $\Sigma_{|i}$. We use the Plackett's differential equation, extended by \cite{berman1987extension}, to find 
\begin{align}
\deriv{}{\Gamma}\Phi_{p-1,\Sigma_{|i}}(-\mm_{|i})\nonumber
&=\nabla\nabla\trans \Phi_{p-1,\Sigma_{|i}}(-\mm_{|i}),
\end{align}
$\nabla\nabla\trans \Phi$ is given in appendix \ref{appendix:CDFhessian}.
\end{itemize}

\section{Generalized \textit{q}-EI as a sum of moments}
\label{appendix:qEI_sum_moments}

\begin{proof}
For given $(\ell,k)$  in $\{1,\ldots,n\}\times\{1,\ldots, q\}$, we consider $E_{\ell,k}$ the event that the random variable inside the expectation term of equation \eqref{eq:qEI:definition} equals $\left(Y(\xx_\ell)-Y(\xx_{n+k})\right)^\alpha$. We have 
\begin{align*}
E_{\ell,k}=\{Y(\xx_{n+k})\le Y(\xx_\ell) \}~&\cap~\left\{\forall i\le n,i\neq\ell; Y(\xx_\ell)\le Y(\xx_i)\right\}\\
&\cap~\left\{\forall j\le q,j\neq k; Y(\xx_{n+k})\le Y(\xx_{n+j})\right\}
\end{align*}

 Considering all pairs $(\ell$, $k$), we have:
\begin{equation}
EI_n(\xx_{n+1:n+q}) = \sum_{\ell=1}^n\sum_{k=1}^{q}\esp_n \left( \left(Y(\xx_\ell)-Y(\xx_{n+k})\right)^\alpha
\indic{E_{\ell,k}}\right)\nonumber.
\end{equation}
For each term $(\ell, k)$ of the sum, the conditioning event can be rewritten $E_{\ell,k}=\{\ZZ^{(\ell,k)}(\xx_{n+1:n+q})\le \zzero\}$, with $\ZZ^{(\ell,k)}$ a random vector of size $n+q-1$, defined by the following linear transformation of $\YY=\left(Y(\xx_{1}),\ldots,Y(\xx_{n+q})\right)\trans$ :

$\forall i=1,\ldots,n+q-1, Z^{(\ell,k)}_i=\left\{\begin{matrix}
\\
\\
\\
\\
\end{matrix}\right.$
\begin{tabular}{ll}
$Y_\ell-Y_i$&$\text{if }1\le i\le \ell-1$ \\
$Y_\ell-Y_{i+1}$&$\text{if } \ell\le i\le n-1$ \\
$Y_{n+k}-Y_{i+1}$&$\text{if } n\le i\le n+q-1,i\neq n+k-1$\\
$Y_{n+k}-Y_{\ell}$&$\text{if }  i =n+k-1$
\end{tabular}
Indeed, the first $n-1$ components of $\ZZ^{(l,k)}\le \zzero $ reflect $\left\{\forall i\le n,i\neq\ell; Y(\xx_\ell)\le Y(\xx_i)\right\}$, and the last components reflect $\left\{\forall j\le q,j\neq k; Y(\xx_{n+k})\le Y(\xx_{n+j})\right\}$ and $\{Y(\xx_{n+k})\le Y(\xx_\ell) \}$.
\end{proof}

\section{Mean square differentiability of $Y(x)^\alpha\indic{B}$}
\label{appendix:meanSquaredDifferentiability}

Let $B$ be an event, $Y$ be a mean-squared differentiable Gaussian process and $\alpha\in\Nset$. Then we have:
\begin{align*}
\esp\left(\left(\frac{Y(x+h)^\alpha-Y(x)^\alpha}{h}\indic{B}-\frac{\drm Y^\alpha}{\drm x}(x)\indic{B}\right)^2\right)\\ \le\esp\left(\left(\frac{Y(x+h)^\alpha-Y(x)^\alpha}{h}-\frac{\drm Y^\alpha}{\drm x}(x)\right)^2\right)\underset{h\rightarrow 0}{\longrightarrow} 0\nonumber
\end{align*} by mean-squared differentiability of $Y^\alpha$.

\section{Symmetry argument}
\label{sec:symmetry}

The term $\frac{q(q+1)}{2}$ comes from a symmetry occurring when summing terms with different index but actually equal.
At fixed summation index $\ell$ in \eqref{eq:qEItransformed2}, we denote  $\omega_{ki}$ the  $i^\text{th}$ term in the scalar product in \eqref{eq:tallisOrdre1} for each $\Mcal_{m+k-1,1}$ required for $\qEI$: $$\forall i,k=1,\ldots,q,~\omega_{ki} =  \Sigma^{(\ell,k)}_{ki}\left[\nabla\Phi_{p, \Sigma^{(\ell,k)}}(-\mm^{(\ell,k)})\right]_i.$$ Then the following symmetry between indices $i$ and $k$ occurs: 
$$\forall i,k=1,\ldots,q,~ \frac{\omega_{ki}}{\Sigma^{(\ell,k)}_{ki}\varphi_{1,\Sigma^{(\ell,k)}_{ii}}(-m^{(\ell,k)}_i)}= \frac{\omega_{ik}}{\Sigma^{(\ell,i)}_{ik}\varphi_{1,\Sigma^{(\ell,i)}_{kk}}(-m^{(\ell,i)}_k))}  $$
Indeed, using the formula of the derivative of CDF, (appendix \ref{appendix:CDFgradient}), leads to:
\begin{align*}
\frac{\omega_{ki}}{\Sigma^{(\ell,k)}_{ki}\varphi_{\Sigma^{(\ell,k)}_{ii}}(-m^{(\ell,k)}_i))}&= \Phi_{p-1,\Sigma^{(\ell,k)}_{|i}}(-\mm^{(\ell,k)}_{|i})\nonumber\\
&= \prob\left(\left. \begin{matrix}\footnotesize Y(\xx_{n+k})\le Y(\xx_{\ell}),\\\footnotesize Y(\xx_{n+j})\le Y(\xx_{n+k}),\forall j=1\ldots q,j\neq k,j\neq i\end{matrix}\right|\begin{matrix}Y(\xx_{n+i})=\\Y(\xx_{n+k})\end{matrix}\right),\nonumber\\
\end{align*}

which is clearly symmetrical between $i$ and $k$.

%\bibliography{article}

\end{document}